\documentclass[journal,twoside,web]{ieeecolor}
\usepackage{tmi}
\usepackage{cite}
\usepackage{amsmath,amssymb,amsfonts}
\usepackage{algorithm}
\usepackage{algorithmic}
\usepackage{graphicx}
\usepackage{textcomp}
\usepackage{times}  
\usepackage{helvet} 
\usepackage{courier}  
\usepackage[hyphens]{url}  
\usepackage{caption} 
\usepackage[font=normal]{subfig}
\usepackage[left]{lineno}
\usepackage{mathrsfs}
\usepackage{multicol}
\usepackage{multirow}
\usepackage{setspace}
\usepackage{booktabs}
\graphicspath{{./images/}}
\newcommand{\tabincell}[2]{\begin{tabular}{@{}#1@{}}#2\end{tabular}}
\makeatletter
\newcommand{\shorteq}{%
	\settowidth{\@tempdima}{-}
	\resizebox{\@tempdima}{\height}{=}%
}
\makeatother
\def\BibTeX{{\rm B\kern-.05em{\sc i\kern-.025em b}\kern-.08em
    T\kern-.1667em\lower.7ex\hbox{E}\kern-.125emX}}
\begin{document}
\title{MixSearch: Searching for Domain Generalized Medical Image Segmentation Architectures}
\author{Luyan Liu, Zhiwei Wen, Songwei Liu, Hong-Yu Zhou, Hongwei Zhu, Weicheng Xie \IEEEmembership{Member, IEEE}, Linlin Shen, \IEEEmembership{Member, IEEE}, Kai Ma, and Yefeng Zheng, \IEEEmembership{Senior Member, IEEE}
\thanks{This work was supported by the grants from Key Area Research and Development Program of Guangdong Province, China (No. 2018B010111001) and the Science and Technology Program of Shenzhen, China (No. ZDSYS201802021814180). Zhiwei Wen and Songwei Liu contribute to this work when they are interns in Tencent Jarvis Lab.}
\thanks{Zhiwei Wen, Weicheng Xie and Linlin Shen are with the College of Computer Science and Software Engineering, Shenzhen 518052, China (email: 1810272035@email.szu.edu.cn; wcxie@szu.edu.cn; llshen@szu.edu.cn)}
\thanks{Luyan Liu, Kai Ma and Yefeng Zheng are with the Tencent Jarvis Lab, Shenzhen 518040, China (e-mail: lly2111101@163.com; kylekma@tencent.com; yefengzheng@tencent.com). } 
\thanks{Songwei Liu and Hongwei Zhu are with the College of Information Science and Electronic Engineering, Zhejiang University, Zhejiang 310058, China (e-mail: 21831068@zju.edu.cn; zhuhw@zju.edu.cn).}
\thanks{Hongyu Zhou is with the Department of Computer Science, University of Hong Kong (e-mail: whuzhouhongyu@gmail.com).}}

\maketitle

\begin{abstract}
Considering the scarcity of medical data, most datasets in medical image analysis are an order of magnitude smaller than those of natural images. However, most Network Architecture Search (NAS) approaches in medical images focused on specific datasets and did not take into account the generalization ability of the learned architectures on unseen datasets as well as different domains. In this paper, we address this point by proposing to search for generalizable U-shape architectures on a composited dataset that mixes medical images from multiple segmentation tasks and domains creatively, which is named ``MixSearch". Specifically, we propose a novel approach to mix multiple small-scale datasets from multiple domains and segmentation tasks to produce a large-scale dataset. Then, a novel weaved encoder-decoder structure is designed to search for a generalized segmentation network in both cell-level and network-level. The network produced by the proposed MixSearch framework achieves state-of-the-art results compared with advanced encoder-decoder networks across various datasets. 
Moreover, we also evaluate the learned network architectures on three additional datasets, which are unseen in the searching process. Extensive experiments show that the architectures automatically learned by our proposed MixSearch surpass U-Net and its variants by a significant margin, verifying the generalization ability and practicability of our proposed method. We make our code and learned network architectures available at: https://github.com/lswzjuer/NAS-WDAN/.
\end{abstract}


\section{Introduction}
\label{sec:introduction}
\IEEEPARstart{T}{he} rise of deep learning~\cite{lecun2015deep} heavily relies on a large amount of labeled data because deep neural networks are able to learn essential feature representations from large-scale datasets. Besides, large-scale data can also help to mitigate the overfitting problem and improve model's ability to generalize to real world applications. However, unlike natural images, which are easy to access and annotate, it is hardly possible to build a large dataset covering common medical image modalities, e.g., X-ray, Computed Tomography (CT) and Magnetic Resonance Imaging (MRI) scans, considering their scarcity and privacy. Consequently, in many cases of medical image segmentation, deep learning methods suffer from the data scarcity problem and the main challenges lie in the following two aspects: 1) limited number of training samples and 2) large variations within multi-domain data. As a result, when coming to real-world applications, well-trained deep models may not be able to produce accurate predictions when tested on a new dataset. To tackle these issues, lots of studies about generalizable deep learning-based methods like domain generalization~\cite{dou2018unsupervised}, transfer learning~\cite{raghu2019transfusion} and self-supervised learning~\cite{zhou2020C2L} have been proposed, recently. However, most of them engineered their task-specific networks to achieve good performance, which triggers our curiosity if there exists a universal network architecture that is suitable for multi-domain, multi-task and small-scale datasets simultaneously. 

\begin{figure}[t]
	\centering
	\subfloat[DedicatedSearch]{\includegraphics[width=0.35\columnwidth]{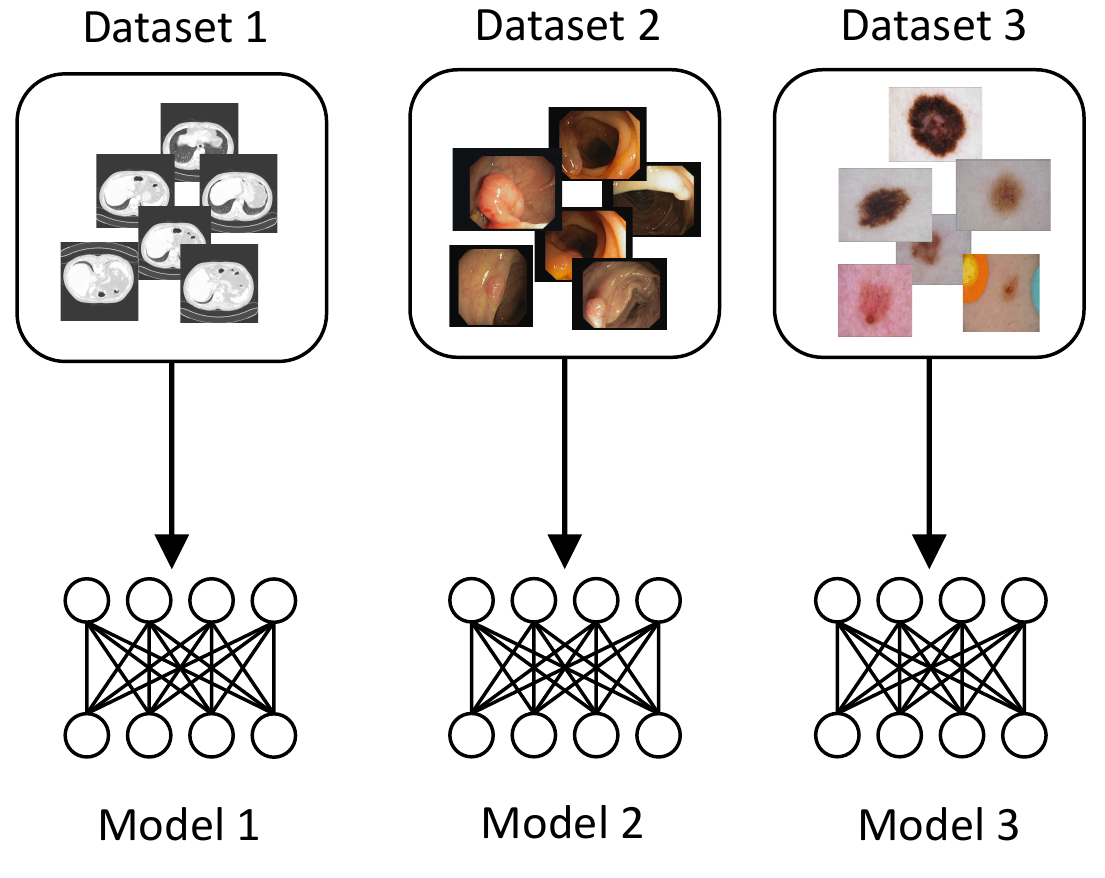}\label{norm_search}}
	\quad
	\subfloat[UnionSearch]{\includegraphics[width=0.3\columnwidth]{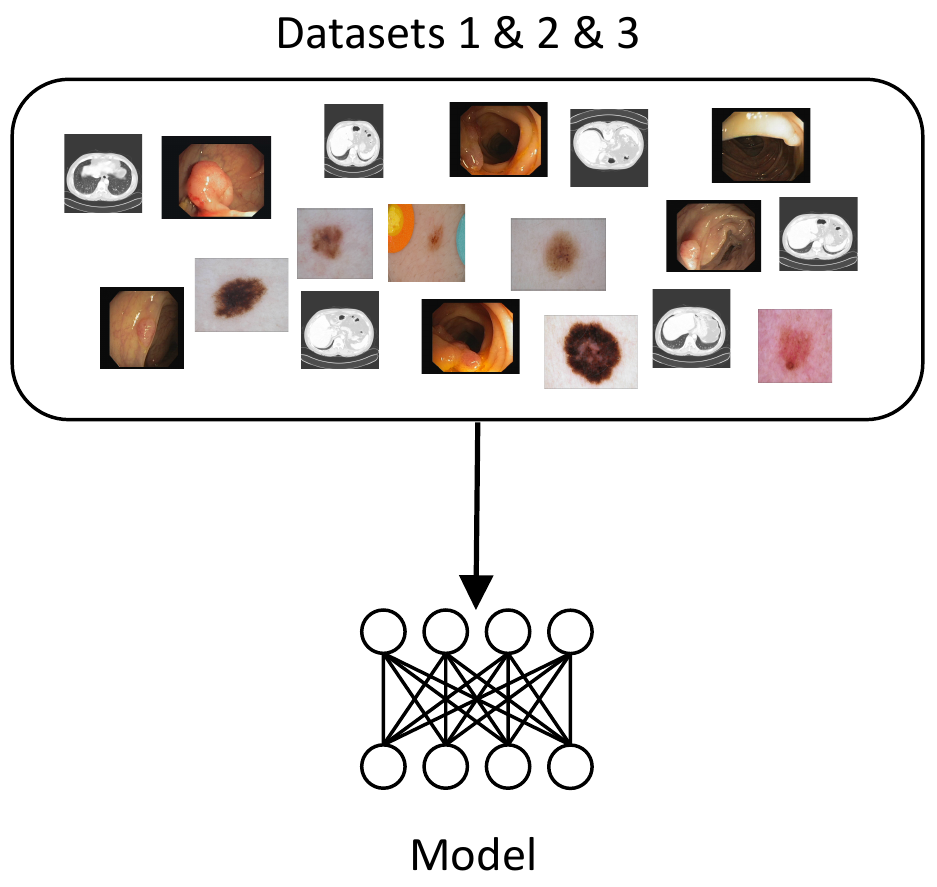}\label{shuffle_search}}
	\quad
	\subfloat[MixSearch]{\includegraphics[width=0.25\columnwidth]{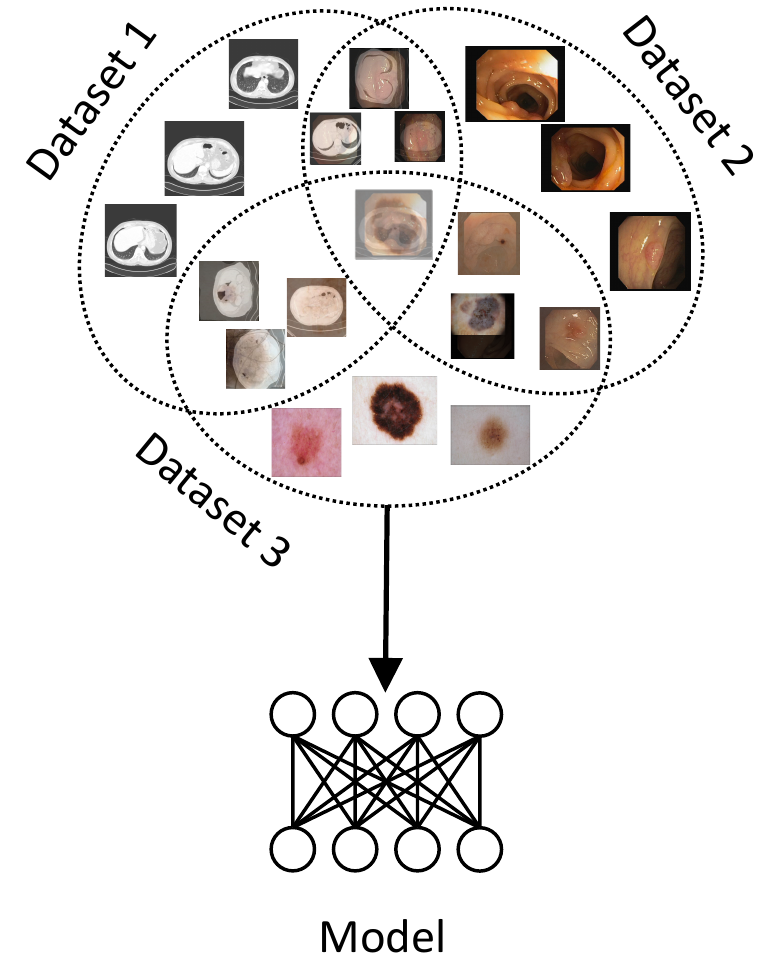}\label{mix_search}}
	\caption{A comparison of different search strategies.} 
	\label{fig:domain_mixup}
\end{figure}

Convolutional Neural Networks (CNNs) have achieved great success in various computer vision tasks based on different hand-crafted network architectures, including image classification~\cite{imagenet,simonyan2014very,resnet}, face recognition~\cite{facenet}, object detection~\cite{ren2015faster,yolo} and image segmentation~\cite{fcn}. Similarly, in the field of medical image segmentation, remarkable progress has also been achieved thanks to the development of CNNs. For example, U-Net~\cite{unet}, as one of the most popular medical image segmentation networks, contains an encoder-decoder architecture and a number of skip connections. Such encoder-decoder structure enables the network to construct powerful image representations by learning to interpret semantics from input images. Meanwhile, skip connections between the encoder layer and corresponding decoder layer can help to incorporate shallow layer features into those deep ones and reduce the difficulty of optimization. Nowadays, there are a lot of U-Net variants addressing the problem of multi-layer fusion, such as Residual U-Net~\cite{drozdzal2016importance}, U-Net++~\cite{unet++} and U-Net3+~\cite{huang2020unet}. However, the classical U-Net and its variants still have a major limitation: most of their skip connections directly connect layers at the same resolution, and do not fully exploit the intermediate representations across different resolutions. Moreover, these networks are designed by experts with their domain knowledge in a long trial-and-error process (usually in months or even years). This process may raise two typical problems: 1) it is hardly possible to manually cover all possibilities in the network space and 2) such trial-and-error process is time-consuming and labor-intensive for human beings and requires a great amount of prior knowledge. 

Nowadays, Neural Architecture Search (NAS) has been developed to search for network architectures automatically. However, traditional NAS requires large-scale datasets with manual annotations which are often unavailable for medical data. In this paper, we propose a novel NAS pipeline, i.e., ``MixSearch", to search for an optimal network architecture on a composited dataset specifically designed for medical image segmentation. The composited dataset includes two parts: 1) original samples from original datasets and 2) composited samples, which are constructed by efficiently mixing samples from different datasets. We argue that the composited dataset has the ability to enlarge the support of the training distribution (i.e., weighted average of the selected images). Moreover, we present a novel search space to search for a better U-shape architecture for medical image segmentation, which is designed as a weaved encoder-decoder structure to break the major limitation of existing encoder-decoder networks. To be specific, inspired by~\cite{aggregating,fabrics}, we develop a novel nested architecture to search for appropriate skip connections, which can explore full-scale feature fusion in top-down, bottom-up and horizontal directions to capture diversified intermediate feature representations. In a nutshell, our main contributions can be summarized into three aspects:

\begin{itemize}
	\item We propose a MixSearch framework to conduct NAS on a composited multi-domain multi-task dataset by mixing multiple small-scale medical datasets. We perform comprehensive studies to demonstrate the strength of using a composited dataset to search for a generalizable deep neural network.
	\item We propose a novel weaved U-shape search space, which enables differentiable cell-level and network-level search to learn optimal feature representations aggregated from different scales and levels. 
	\item Experimental results demonstrate that our proposed MixSearch approach is able to produce more generalizable network architecture, which outperforms existing state of the art by a significant margin. It is worth noting that the model learned by MixSearch achieves the best results on both seen and unseen datasets.
\end{itemize}

\section{Related Works}
\subsubsection{Medical Image Segmentation} Medical image segmentation is a critical step in medical image analysis. Initially, it was done with sequential application of low-level pixel processing. With the development of deep learning techniques, medical image segmentation methods are currently dominated by CNNs. Ronneberger \emph{et al.} \cite{unet} took the idea of Fully Convolutional Networks (FCNs) \cite{fcn} one step further and proposed an encoder-decoder architecture named U-Net, which has become one of the most commonly used benchmark models for medical image segmentation. Recently, there are many U-Net variants proposed for further improvements. Oktay \emph{et al.} \cite{attention_unet} proposed Attention U-Net, which incorporates attention mechanism into classical U-Net by automatically focusing on target structures of varying shapes and sizes. Noticing the limitations of skip connections, Zhou \emph{et al.} \cite{unet++} proposed U-Net++ to redesign the way of implementing skip connections and introduced a built-in ensemble of U-Nets with varying depths. MultiResU-Net \cite{multiresunet} replaced skip connections with residual paths to reduce the semantic gap between the corresponding levels of encoders and decoders. However, all of them still followed a hand-crafted manner to design new architectures, which is hard to search all the possibilities to tackle the limitations of current encoder-decoder architecture with skip connections. 

\subsubsection{Neural Architecture Search in Natural Images} Although deep CNNs have achieved great success in the field of computer vision, designing a good network still heavily relies on expert experience and is time-consuming and laborious. Neural Architecture Search (NAS) aims to automatically design a neural network, which is an emerging topic of AutoML and has attracted increasing attentions from both academia and industry~\cite{mobilenetv3}. According to the heuristics to explore large architecture space, there are three kinds of NAS approaches, i.e., reinforcement learning based approaches~\cite{nasnet,ENAS}, evolution based approaches~\cite{regularized} and differentiable approaches~\cite{darts,pdarts,pcdarts}. Among them, differentiable approaches combining search and validation processes achieve higher search efficiency and better performance in the classification tasks. DARTS ~\cite{darts} was the first differentiable NAS algorithm, which relaxed the search space to be continuous so that the architecture could be optimized by using gradient descent. Besides the classification tasks, NAS has also been used to explore neural network architecture for semantic segmentation. Chen \emph{et al.} \cite{dpc} was dedicated to search a multi-scale cell with the efficient random search. Zhang \emph{et al.} \cite{customizable} used differentiable NAS to search a normal cell, a reduce cell and a multi-scale cell. These cells were stacked into a lightweight network. Liu \emph{et al.} \cite{auto} constructed a hierarchical search space to handle the spatial resolution changes. Although the above studies utilized NAS to improve the segmentation accuracy, all of them were oriented to natural images. 

\subsubsection{Neural Architecture Search in Medical Images} In medical image analysis, Kim \emph{et al.} \cite{kim2019scalable}, Zhu \emph{et al.} \cite{zhu2019v} and Yu \emph{et al.} \cite{yu2020c2fnas} applied NAS to search efficient 3D volumetric medical segmentation networks. Weng \emph{et al.} \cite{nas-unet} directly used the differential NAS to search for optimal cell-level operations and simply stacked the same number of downsampling and upsampling cells to construct a U-Net like architecture. However, the above works ignored a typical problem, i.e., multi-domain small-scale datasets, in medical images. Most of them searched networks on a designated dataset, which sets limits to the generalization ability of the learned architectures. In this paper, we address this issue by proposing a mixing strategy to build a composited dataset based on multiple small-scale datasets from different domains. Moreover, we propose to evaluate the generalization ability of learned architectures on both seen and unseen datasets. We also improve the searching strategy where the search space is redesigned as a novel nested/weaved encoder-decoder structure with a multi-layer feature fusion branch. In practice, both cell-level and network-level infrastructures can be learned simultaneously, resulting in an efficient search process. 

\begin{figure}[t]
	\centering
	\includegraphics[width=1\columnwidth]{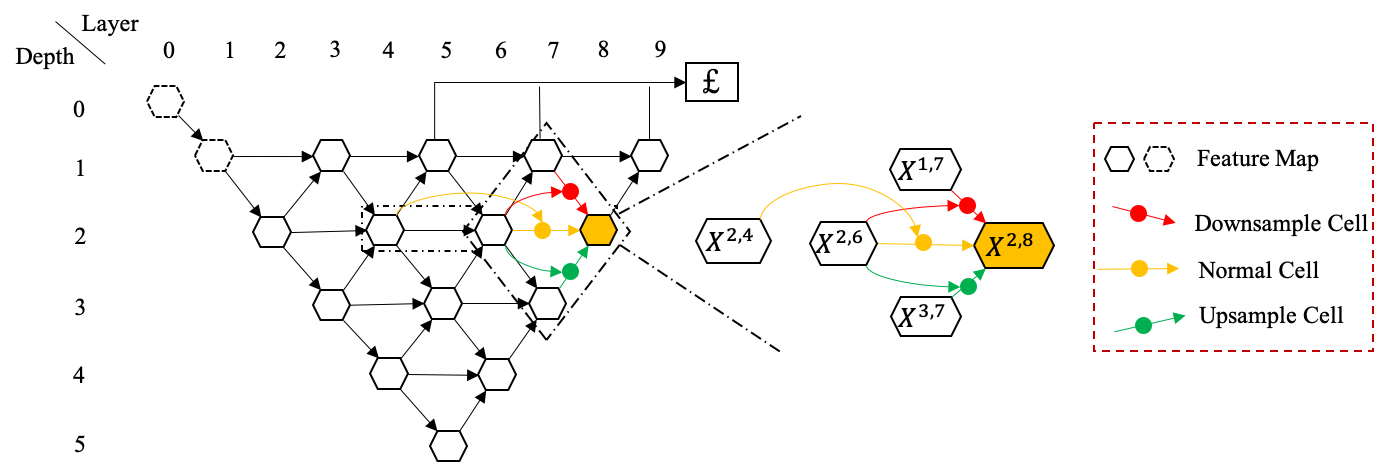}
	\caption{An overview of the proposed search strategy.} 
	\label{cell_out}
\end{figure}

\begin{figure}[t]
	\centering
	\subfloat[]{\includegraphics[width=0.26\columnwidth]{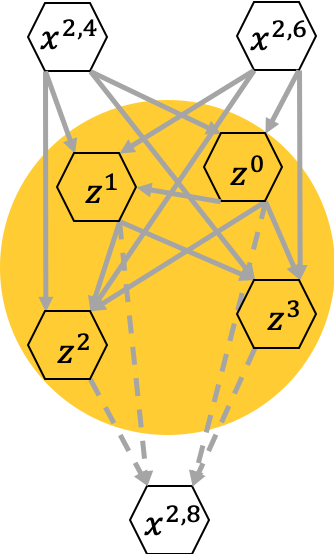}\label{cell_in}}
	\quad
	\subfloat[]{\includegraphics[width=0.4\columnwidth]{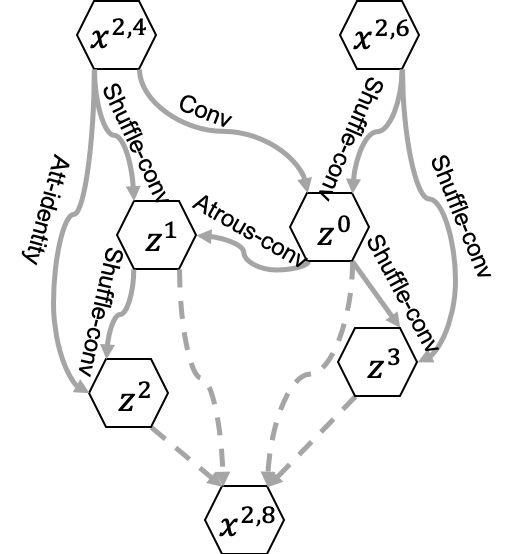}\label{cell_in_res}}
	\caption{(a) The cell-level search graph. (b) An example of normal cell architecture learned by MixSearch.} 
	\label{fig:cell_in_fig}
\end{figure}

\section{Proposed MixSearch}
In this section, we firstly introduce how to build a composited dataset across different domains and tasks. Then, we bring in the novel designed architecture search strategy, which helps to produce a better encoder-decoder architecture by fully exploring its parametric space.

\subsection{Dataset Mixing}
\label{3.1}
The privacy and scarcity of medical data prevent the community from building large-scale datasets as those of natural images. Even if we can have access to some massive data, it is still difficult to obtain manual annotations from clinical experts. On the other hand, the clinical data collected from various medical centers or equipments usually contain domain shift. Thus, searching neural networks on only one dataset cannot guarantee satisfying performance when tested on other datasets. In this paper, we propose to mix those medical image datasets from various domains and incorporate them into a composited one. Such dataset not only includes the real images from different segmentation datasets, but also contains synthetic (or virtual) data generated by mixing different datasets. In practice, suppose we currently have $N$ medical datasets with manual annotations, we can denote each dataset by $\mathcal{D}_{i}$, where $i\in \{1,2,...,N\}$. Obviously, we have two strategies to construct target datasets for the need of performing NAS. Firstly, we can simply search for a specific neural network using a specific $\mathcal{D}_{i}$ only (cf. DedicatedSearch in Fig. 1). However, such strategy will undoubtedly lead to two results: 1) the search process tends to overfit this small-scale dataset easily and 2) the generalization ability of the learned architecture on other datasets cannot be guaranteed. Besides using $\mathcal{D}_{i}$ only, we can choose to combine different datasets by simply uniting them (cf. UnionSearch in Fig. 1):
\begin{linenomath*}
	\begin{equation}
	\mathcal{D}_{\rm{U}}=\mathcal{D}_{\rm{1}}\cup \mathcal{D}_{\rm{2}} \cup ... \cup \mathcal{D}_{\rm{N}}.
	\end{equation}
\end{linenomath*}
Although such intuitive strategy may be helpful in some cases, it cannot increase the number of potential training samples efficiently and exponentially. 

Considering these reasons, we propose to build a composited dataset by applying an idea similar to mixup~\cite{zhang2017mixup} to expand the distribution of $\mathcal{D}_{\rm{U}}$. As far as we know, mixup regularizes the neural network to favor simple linear behavior in-between training examples by training a neural network on convex combinations of pairs of examples and their labels sampled from a \emph{single dataset}. In this paper, we further extend it to the dataset level where the composited dataset is constructed using the weighted sum of several images randomly sampled from different datasets. We extends traditional mixup, which is simply based on two randomly sampled instances, to a multi-instance version. Specifically, $k$ is a predefined hyperparameter and we randomly sample $k$ different samples from $D_{\rm{U}}$ and then perform sample mixing on top of them. It is worth noting that we propose to increase $k$ to be larger than two (the value in mixup), which has shown to be able to consistently gain performance. According to the rule of mixup, a composited sample $\tilde{x_l}$ can be formulated as:\footnote{Counter-intuitively, such an image is unreadable to humans but can improve the robustness of the network trained on it.}
\begin{linenomath*}
	\begin{equation}
	\label{mixup1}
	\tilde{x}_l=\tilde{\lambda}_1 x_1+\tilde{\lambda}_2 x_2+...+\tilde{\lambda}_k x_{k}
	\end{equation}
\end{linenomath*}
where
\begin{linenomath*}
	\begin{equation}
	\label{vrm1}
	\tilde{\lambda}_1, \tilde{\lambda}_2, ..., \tilde{\lambda}_k = \rm{softmax}(\lambda_1, \lambda_2, ..., \lambda_k),
	\end{equation}
\end{linenomath*}
where $\{\lambda_1, \lambda_2, ..., \lambda_k\}$ are drawn from a Beta distribution, i.e., $\rm{Beta}(\mu,\mu)$, independently, and hyperparameter $\mu \in(0,+\infty)$. After softmax normalization, $\{\tilde{\lambda}_1, \tilde{\lambda}_2, ..., \tilde{\lambda}_k\} \in[0,1]$ and their sum equals one.
Accordingly, the label of input $\tilde{x_l}$ is:
\begin{linenomath*}
	\begin{equation}
	\label{mixup2}
	\tilde{y}_l=\tilde{\lambda}_1 y_1+\tilde{\lambda}_2 y_2+...+\tilde{\lambda}_k y_{k},
	\end{equation}
\end{linenomath*}
where $\{$($x_{1},  y_{1}$), ($x_{2}$, $y_{2}$), ..., ($x_{k}$, $y_{k}$)$\}$ are instance-label vectors drawn from the composited training data $\mathcal{D}_{\rm{U}}$, and $\{y_1, y_2, ..., y_k\}$ are one-hot label encodings. The composited dataset can be denoted as $\mathcal{D}_{\rm{M}}= \{(\tilde{x_l},\ \tilde{y_l})\}_{1}^{m}$ which includes the additional synthetic samples drawn from the vicinity distributions of different datasets, and $m$ is the total number of the virtual samples generated from $k$ different datasets. The prior knowledge leveraged in Eqs. (\ref{mixup1}) and (\ref{mixup2}) is that the linear combination of training images would lead to the linear combination of corresponding labels, which would in turn extend the distribution of training data. Such virtual composited dataset would encourage the learned network $\mathcal{F}$ to behave linearly in-between training samples from one dataset to the other. The optimization problem of minimizing the average of the loss function $\mathcal{L}$ over the composited data distribution $P$ can be solved by minimizing the empirical risk: 
\begin{linenomath*}
	\begin{equation}
	\label{minize}
	R_{v}=\frac{1}{m}\sum_{l = 1}^{m}\mathcal{L}(\mathcal{F}(\tilde{x}_l),\ \tilde{y}_l),
	\end{equation}
\end{linenomath*}
where $(\tilde{x}_l,\ \tilde{y}_l) \in D_{\rm{M}}$. 

In summary, the proposed mixing strategy has several advantages besides exponentially increasing the number of training samples (a composited dataset with an infinite number of samples can be constructed). On the one hand, when the sampled images belong to the same dataset, $(\tilde{x}_l,\ \tilde{y}_l)$ enlarges the support of this intra-dataset training distribution, which is in accordance with the explanation in \cite{zhang2017mixup}. Searching neural networks on this expanded single domain dataset can reduce the risk of overfitting and improve the performance of learned model. On the other side, when the sampled images belong to different datasets, $(\tilde{x}_l,\ \tilde{y}_l)$ provides smooth transition between different domains (inter-dataset distribution). By enforcing the network to learn representations from composited data during the architecture searching process, our MixSearch approach is able to fit a continuous multi-center distribution and produce a high-performance and generalizable model architecture.

\begin{table}
	\small
	\caption{A summary of candidate operations used in different cells. \emph{Conv.} and \emph{Convt.} stands for convolution and transposed convolution, respectively. \emph{Identity} and \emph{Att-identity} denote residual connection without and with attention mechanism \cite{senet}, respectively. \emph{Sep-conv.}, \emph{Atrous-conv.} and \emph{Shuffle-conv.} correspond to the depth separable convolution \cite{mobilenets}, atrous convolution \cite{deeplabv3} and group-shuffle convolution \cite{shufflenet}, respectively. The number enclosed in the parentheses denotes the operational stride.}
	\begin{tabular}{lll}
		\toprule
		Down-ops      & Up-ops       & Normal-ops  \\
		\midrule
		Average Pool. (2) & Convt. (2)    & Identity \\
		Max Pool. (2)      & Atrous-convt. (2) & Att-identity \\
		Conv. (2)       & Sep-convt. (2)       & Conv. (1)\\
		Atrous-conv. (2)   & Att-convt. (2)   & Atrous-conv. (1) \\
		Sep-conv. (2)     & -    & Sep-conv. (1) \\
		Att-conv. (2)  & -              & Shuffle-conv. (1) \\
		\bottomrule
	\end{tabular}
	\label{ops}
\end{table}

\subsection{Searching in Weaved Space}

We notice that the skip connections in U-Net and its variants may not be able to fully explore the potential of multi-scale fusion. Specifically, traditional U-Net only merges feature maps that have the same resolution, while ignoring intermediate semantics. In fact, U-Net++ \cite{zhou2018unet++} addressed this problem by proposing to use an intuitive nested architecture where they simply used a convolutional layer to deal with concatenated representations. In this paper, we argue that the above implementation has not paid enough attention on how to merge multi-layer outputs efficiently, and propose to use a novel weaved architecture searching method to tackle this problem.

In order to break the limitations of fixed skip connections, we decide to search for an optimal path. We provide an overview of the proposed search strategy in Fig. \ref{cell_out}. In general, we initialize the search space as a weaved acyclic graph. The depth (D) and layer (L) in Fig. \ref{cell_out} represent the index of each feature map in vertical and horizontal directions, respectively. Each feature map in the network is represented by a hexagon where $X^{d,l}$ stands for the feature map lying in depth $d$ and layer $l$. Each feature map in Fig. \ref{cell_out} shares the same connection structure as $x^{2,8}$ (which is highlighted in a zoomed-in view); however, we do not show all of them for brevity. More details will be introduced in the following based on Fig. \ref{cell_out}.

\subsection{Optimization Strategies}
\label{optim}


In practice, we perform cell-level search and network-level search simultaneously. The cell-level search is defined as conducting gradient optimization within each cell (colored circles in Fig. 2) to select the optimal combination of different operations (cf. Table \ref{ops} and Table \ref{sets}). The network-level search aims at searching for the optimal combination of different cells (e.g., the colored lines in Fig. \ref{cell_out}). Following DARTS \cite{darts}, we use continuous relaxation to turn the problem of NAS into a gradient-based super-network optimization process. Furthermore, we design a self-contained weaved search space to search for an optimal network with ample feature aggregation.

\begin{table}[htp]
	\small
	\caption{Different cells employ different sets of operations.}
	\centering
	\begin{tabular}{cccc}
		\toprule
		Cell type  & Down-ops ($O_{1}$) & Normal-ops ($O_{2}$) & Up-ops ($O_{3}$) \\
		\midrule
		Downsample & \checkmark & \checkmark & \\
		Normal & & \checkmark & \\
		Upsample & & \checkmark & \checkmark\\
		\bottomrule
	\end{tabular}
	\label{sets}
\end{table}

\begin{algorithm}
	\caption{Two stage differentiable neural architecture search.}
	\small{\textbf{Input}: Composited dataset $D_{mixup}$, which is divided into weight-training set and architecture-training set; candidate operations sets $O_1$, $O_2$ and $O_3$; learning rate $\eta_\alpha$, $\eta_\beta$ and $\eta_w$; search epochs $K$ and warm-up epoch $k_{warm\_up}$; network scale parameters  $D_{0}$, $D_{1}$, $L_{0}$, $L_{1}$.}
	
	\small{\textbf{Output}: Architecture parameters $\alpha$ and $\beta$, which represent the final cell and network structure, respectively.}
	\label{alg1}
	\begin{algorithmic}[1]
		
		\STATE Two stage searching begins from $s = 0, k=0$
		\WHILE{$s<2$}
		\STATE Build super-network $f(w;\alpha,\beta)$
		\STATE Initialize $w$, $\alpha$, $\beta$, based on $D_{s}$, $L_{s}$, $O_1$, $O_2$ and $O_3$
		\WHILE{$k\le K$}
		\IF{$k<k_{warm\_up}$}
		\STATE   $w_{k+1}\leftarrow w_{k}-\eta_w\cdot\bigtriangledown_{w_{k}}\mathcal{L}_{Wei}(w_k;\alpha_k,\beta_k)$
		\STATE   $\beta_{k+1}\leftarrow \beta_{k}$
		\STATE   $\alpha_{k+1}\leftarrow \alpha_{k}$
		\ELSE
		\STATE   $w_{k+1}\leftarrow w_{k}-\eta_w\cdot\bigtriangledown_{w_{k}}\mathcal{L}_{Wei}(w_k;\alpha_k,\beta_k)$
		\STATE   $\beta_{k+1}\leftarrow \beta_{k}-\eta_\beta\cdot\bigtriangledown_{\beta_{k}}\mathcal{L}_{Arch}(w_{k+1};\alpha_k,\beta_k)$
		\STATE   $\alpha_{k+1}\leftarrow \alpha_{k}-\eta_\alpha\cdot\bigtriangledown_{\alpha_{k}}\mathcal{L}_{Arch}(w_{k+1};\alpha_k, \beta_{k+1})$
		\ENDIF
		\STATE   $k\leftarrow {k+1}$
		\ENDWHILE
		\STATE   $s\leftarrow {s+1}$
		\STATE   $O_1\leftarrow {halve(O_1)}$ \label{halve1}
		\STATE   $O_2\leftarrow {halve(O_2)}$ \label{halve2}
		\STATE   $O_3\leftarrow {halve(O_3)}$ \label{halve3}
		\ENDWHILE
		\RETURN $\alpha\leftarrow\alpha_K$,  $\beta\leftarrow\beta_K$
	\end{algorithmic}	
\end{algorithm}

\subsubsection{Cell-level Search} As shown in Fig. 2 and Table \ref{sets}, we design three types of cell architectures, named Normal cell, Downsample cell and Upsample cell, which represent horizontal, top-down, and bottom-up feature fusion, respectively. We introduce the three types of cell architectures as follows. The resolution of two input feature maps of Normal cell is the same with that of the output feature map; therefore, Normal cell can aggregate the horizontal level feature maps iteratively and create rich intermediate feature representations to reduce the semantic gaps between the same-scale feature maps in the encoder-decoder networks. One of the input feature map in Downsample cell is from a shallower layer, which contains more detailed spatial information, thus enables the Downsample cell to enhance the feature representation globally. One of the input feature map in Upsample cell is from a deeper layer, which captures higher order and abstract semantic information. The feature maps from different depths and scales can be fused together and enrich the semantic feature representations.

To introduce the cell-level search strategy more clearly, we take the Normal cell in Fig. \ref{cell_out} (cf. the yellow circle in Fig. \ref{cell_in}) as an example. As shown in Fig. \ref{cell_out}, a searchable cell is represented as a Directed Acyclic Graph (DAG) with four intermediate feature maps (i.e., $\{z^0, z^1, z^2, z^{3}\}$), two inputs (i.e., $x^{2,4}$ and $x^{2,6}$) and one output (i.e., $x^{2,8}$) feature maps. The other two type of cells (i.e., Downsample cell and Upsample cell) share the same design as the Normal cell. 
The constructed set of candidate operations is $O$ and each element in $O$ is defined as $o(\cdot)$. The information flow connecting any two feature maps $i$ and $j$ can be represented using an edge $E(i, j)$, which consists of a set of operations weighted by the parameters $\alpha^{(i, j)}$ of dimension $|O|$. Then,  we can apply softmax to all possible operations:
\begin{linenomath*}
	\begin{equation}
	\label{eq:soft}
	\tilde{o}^{(i, j)}(x^i)=\sum_{o\in O}\frac{exp(\alpha^{(i, j)}_o)}  {\sum_{o^{\prime}\in O}exp(\alpha^{(i, j)}_{o^{\prime}})}o(x^i),
	\end{equation}
\end{linenomath*}
where $\tilde{o}^{(i, j)}$ stands for the weighted operations. An intermediate feature map can be represented as $x^j=\sum_{i \textless j}\tilde{o}^{(i, j)}(x^i)$. Finally, the output latent representation $x^{2,8}$ in Fig. \ref{cell_in} is produced by concatenating $z^0$, $z^1$, $z^2$ and $z^3$. At the end of network search, the operation with the highest probability in each mixed-operation is retained, which can be formulated as $\tilde{o}^{(i, j )}(x^i)=\mathop{\arg\max}_{o\in O} \alpha^{(i, j)}_o$. 


\subsubsection{Network-level Search} As shown in Fig. \ref{cell_out}, our weaved architecture gradually aggregates and refines different feature maps from different layers. Each intermediate feature map in the network contains top-down, horizontal and bottom-up operations to equalize resolution and standardize semantics. In order to study the effectiveness of each feature fusion branch and guide the final pruning process, we introduce continuous relaxation, which means that each feature map $X^{d, l}$ can be represented as a weighted sum of the outputs of three cells:
\begin{linenomath*}
	\begin{align}
	\label{eq:relaxb}
	\notag
	X^{d,l} &= p^{d,l}_0 cell_{\rm{d}}(X^{d,l-2},X^{d-1,l-1})\\
	\notag
	&+p^{d,l}_1 cell_{\rm{n}}(X^{d,l-4},X^{d,l-2})\\
	&+p^{d,l}_2 cell_{\rm{u}}(X^{d,l-2},X^{d+1,l-1})
	\end{align}
\end{linenomath*}
where $p=\rm{softmax}(\beta)$ and $\beta\in \mathbb{R}^{\rm{D}\times \rm{L}\times3}$ is also a set of parameters, which can be optimized using gradient descent. $cell_{\rm{d}}$, $cell_{\rm{n}}$, $cell_{\rm{u}}$ represent Downsample cell, Normal cell and Upsample cell, respectively.

As regard to the complexity analysis, each cell is set to contain four intermediate feature maps ($M=4$), which is similar to the DARTS \cite{darts}, the total number of the learnable edges between all the intermediate and input feature maps is $E=2M+M(M-1)/2=14$.
Let's define the connection between each two feature maps in a searchable cell as an edge. For Downsample cell, there are ten edges constructed by $O_{3}$ and four edges constructed by $O_{1}$. Similarly, there are also ten edges constructed by $O_{3}$ and four edges constructed by $O_{2}$ in Upsample cell. All the 14 edges of the Normal cell form  $O_{3}$. Considering the cell-level search only, the total number of the candidate architectures in the continuous search space before discretization is $7^{10}*(6^{4}+4^{4}+7^{4})\approx2.25\times10^{17}$.
It is greater than that of PNAS \cite{liu2018progressive} searching for one single type of cell, which is $5.6\times10^{14}$. 

\begin{table*}[htp]
	\footnotesize
	\centering
	\caption{Searching segmentation models on ISIC \cite{isic2018}, CVC \cite{cvc-db}, CHAOS-CT \cite{chao2019}, the union set (denoted as \emph{Union}) and our composited dataset (denoted as \emph{Composite}), respectively. \emph{DC} and \emph{DS} denote channel doubling and deep supervision, respectively.}
	\begin{tabular}{l|c|c|c|c|c|c|c|c|c|c|c}
		\toprule
		\multirow{2}{*}{Method} &\multirow{2}{*}{\tabincell{c}{Datasets\\{\footnotesize (source)}}} &\multirow{2}{*}{$\mu$} &\multirow{2}{*}{DC}  &\multirow{2}{*}{DS}
		&\multicolumn{2}{c|}{ISIC (target)} &\multicolumn{2}{c|}{CVC (target)} &\multicolumn{2}{c|}{CHAOS-CT (target)} & Params \\
		\cline{6-11}
		& & & & & Dice (\%) & Jc (\%) & Dice (\%) & Jc (\%) & Dice (\%) & Jc (\%) & (M) \\
		\hline
		\hline
		U-Net \cite{unet} & -  & - &  \checkmark  & & 86.47 & 77.24 & 90.13 & 82.29 & \textbf{94.85} & \textbf{90.30} & 34.53 \\
		U-Net++ \cite{unet++} & - & - & \checkmark  & \checkmark & 86.44 & 77.12 & 90.63 & 82.95 & 94.21 & 89.52 & 36.63 \\
		Att. U-Net \cite{attention_unet} & - & - & \checkmark  & & \textbf{87.22} & \textbf{78.04} & \textbf{91.07} & \textbf{83.49} & 94.41 & 89.82 & 34.89\\
		MR U-Net \cite{multiresunet} & - & - &\checkmark  & & 86.59 & 77.44 & 90.34 & 82.53 & 92.82 & 87.49 & 34.84\\
		R2T U-Net \cite{recurrent} & - & - & \checkmark  & & 85.02 & 75.05 & 89.61 & 81.30 & 92.17 & 85.88 & 39.09\\
		\hline
		\multirow{3}*{DedicatedSearch} & ISIC & - &\checkmark  & \checkmark & \textbf{88.53} & \textbf{79.80} & 91.28 & 84.06 & 92.29 & 85.97 & 8.81 \\
		& CVC & - & \checkmark  & \checkmark & 87.41 & 78.11 & \textbf{91.96} & \textbf{85.24} & \textbf{94.60} & \textbf{89.90} & 10.40 \\
		& CHAOS-CT & - & \checkmark  & \checkmark & 86.30 & 76.48 & 89.60 & 81.28 & 93.62 & 88.22 & 4.36\\
		\hline
		UnionSearch & Union & - & \checkmark  & \checkmark & 88.58 & 79.85 & 91.73 & 84.85 & 94.12 & 89.07 & 9.71\\
		\hline
		MixSearch ($k$\shorteq 2) & CVC & 1 & \checkmark  & \checkmark & 88.80 & 80.87 & 92.11 & 85.43 & 95.02 & 90.74 & 7.67 \\
		MixSearch ($k$\shorteq 3) & CVC & 1 & \checkmark  & \checkmark & 88.05 & 80.02 & 91.71 & 84.77 & 94.83 & 90.37 & 5.70 \\
		MixSearch ($k$\shorteq 4) & CVC & 1 & \checkmark  & \checkmark & 89.03 & 81.15 & 92.24 & 85.72 & 94.05 & 88.98 & 7.37 \\
		\hline
		\multirow{2}*{MixSearch ($k$\shorteq 2)} & Composite& 0.5 & \checkmark & \checkmark & \textbf{88.55} & \textbf{79.86} & \textbf{93.00} & \textbf{87.00} & \textbf{95.79} & \textbf{92.03} & 12.79\\
		& Composite & 1 & \checkmark  & \checkmark & 88.52 & 79.74 & 91.99 & 85.32 & 94.80 & 90.32 & 10.55\\	 
		\hline
		\multirow{2}*{MixSearch ($k$\shorteq 3)} & Composite & 0.5 & \checkmark & \checkmark & 89.07 & 81.25 & 91.53 & 84.58 & 95.06 & 90.51 & 12.19\\
		& Composite & 1 & \checkmark  & \checkmark & \textbf{90.07} & \textbf{82.68} & \textbf{92.75} & \textbf{86.71} & \textbf{95.68} & \textbf{91.76} & 9.54\\
		\hline
		\multirow{2}*{MixSearch ($k$\shorteq 4)} & Composite & 0.5 & \checkmark  & \checkmark & \textbf{89.95}&\textbf{82.47}&\textbf{93.03}&\textbf{87.05}&\textbf{95.21}&\textbf{91.01}&10.98\\
		& Composite & 1 & \checkmark  & \checkmark & 89.63 & 82.17 & 92.06 & 85.41 & 94.79 & 90.34 & 13.26 \\	 
		\bottomrule
	\end{tabular}
	\label{tab:compare}
\end{table*}

\subsubsection{Bilevel optimization} After performing continuous relaxation modeling, the super-network can be written as $f(w;\alpha,\beta)$, where $\alpha$, $\beta$ represent the architecture parameters in cell-level (cf. Eq. (6)) and network-level (cf. Eq. (7)) search, respectively; $w$ denotes the network weights (depending on $\alpha$ and $\beta$). The composited data is split into two parts: weight-training (50\%) and architecture-training (50\%) set. The loss functions are denoted by $\mathcal{L}_{Wei}$ and $\mathcal{L}_{Arch}$, which are determined by the network weights ($w$) and architecture parameters ($\alpha$ and $\beta$), respectively. Our goal is to find optimal $\alpha^{*}$, $\beta^{*}$  that can minimize the architecture loss $\mathcal{L}_{Arch}(w^{*};\alpha^{*},\beta^{*})$:
\begin{linenomath*}
	\begin{align}
	\label{eq:s.t.}
	\notag
	& \min_{\alpha,\beta} \mathcal{L}_{Arch}(w^{*}(\alpha,\beta),\alpha,\beta),\\
	& {\rm{s.t.}}\ w^{*}(\alpha,\beta)=\mathop{\arg\min}_{w} \mathcal{L}_{Wei}(w,\alpha,\beta).
	\end{align}
\end{linenomath*}

Similar to \cite{darts}, we optimize the super-network following a bilevel optimization process with ($\alpha, \beta$) as the upper-level variable and $w$ as the lower-level variable. Before optimizing $\alpha$ and $\beta$, we need to determine current $w^{*}$ by solving Eq. (\ref{eq:s.t.}). But solving this embedded optimization problem requires training the network until convergence, which is very time-consuming and inefficient. So we approximate $w^{*}$ by adapting $w$ using only a single training iteration. In order to ensure a sufficiently large-scale search space, and reduce search time and memory footprint during search process, we adopt a two-stage search strategy ~\cite{pdarts}, which is shown in Algorithm 1. In the first stage, we use three complete sets of candidate operations to search for a shallow network (e.g., $D_0=5, L_0=8$), which is to efficiently evaluate each candidate operation with a lower memory footprint and a faster speed. In the second stage, we remove half of the candidate operations (cf. line \ref{halve1}, \ref{halve2} and \ref{halve3} in Algorithm \ref{alg1}), which contribute the least  as evaluated by the first stage. Then, we use the updated smaller sets of candidate operations to build a deeper network (e.g., $D_1=6, L_1=10$). A total of 80 epochs are used for architecture learning in each stage, in which the first 10 epochs are used for warm-up. The whole search process is described in Algorithm \ref{alg1}.

\section{Experiments}

\subsection{Implementation Details}
The proposed method is implemented using PyTorch on two Tesla P40 GPUs with 48 GB memory in total. A cosine annealed SGD optimizer \cite{bottou2010large} is used to optimize $w$, with an initial learning rate of 0.025. The architectural parameters ($\alpha$, $\beta$) are optimized using the Adam optimizer \cite{adam} with a fixed learning rate of $2 \times 10^{-4}$, a momentum of (0.5, 0.999) and a weight decay of $1 \times 10^{-3}$. The proposed two-stage search algorithm is performed on the composited dataset with 80 epochs. We split each dataset into a training set (90\%)  and a test set (10\%). All training sets are combined to generate a composited dataset (cf. Eqs. (2), (3) and (4)) to learn an optimal architecture with MixSearch. Then, we retrain the learned architecture from scratch on the original training set for each domain and report their performance on the corresponding test set. In the retraining stage, the number of training epochs for each experiment is 1600. We use a cosine annealed SGD optimizer with an initial learning rate of $5 \times 10^{-3}$, a momentum of 0.9, a weight decay of $1 \times 10^{-5}$. The loss function used in the experiments are the weighted sum of cross entropy loss and Dice loss. Our proposed search strategy is efficient because the entire search process takes only 1.9 GPU days (two P40 GPUs) with the batch size of 12. For simplicity, we denote the dataset used for architecture search as the \emph{source} dataset, while the test dataset is identified as the \emph{target} dataset.

\begin{figure}
	\centering
	\includegraphics[width=0.8\columnwidth]{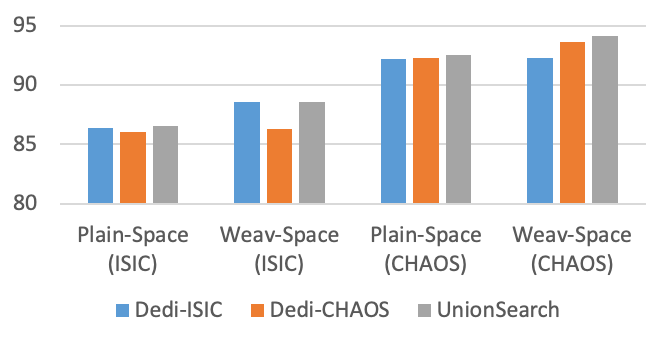}%
	\caption{The comparison of the learned models by NAS-Unet \cite{nas-unet} on the plain space and weaved space.}
	\label{fig:compare with NAS-UNET}
\end{figure}

\begin{figure}
	\centering
	\includegraphics[width=0.8\columnwidth]{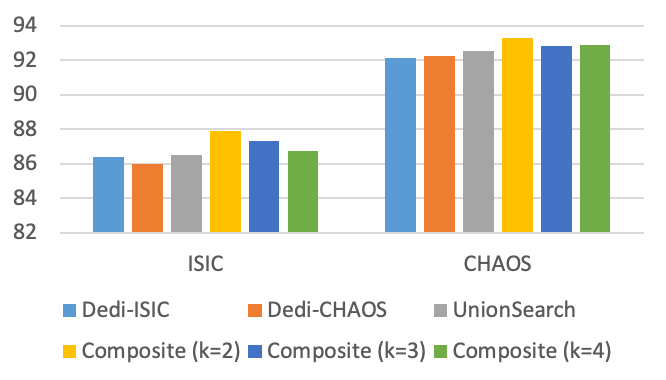}%
	\caption{The performance of NAS-Unet \cite{nas-unet} using the mixed dataset.}
	\label{fig:mixsearch of NAS-UNET}
\end{figure}

The performance of different models generated by various search methods is evaluated using Dice score and Jaccard score~\cite{mcguinness2010comparative} on three datasets, including ISIC~\cite{isic2018}, CVC~\cite{cvc-db} and CHAOS-CT~\cite{chao2019}, which are collected from dermoscopy, gastroscopy and CT equipments, respectively. In order to evaluate the generalization ability of learned architecture across different datasets and domains, we also conduct extensive experiments on three additional datasets (unseen in search stage) which are ETIS \cite{silva2014toward}, KiTS \cite{heller2019state} and LiTS \cite{lits}. 

\begin{table}[!htp]
	\fontsize{8.5}{10}\selectfont
	\centering
	\caption{Effectiveness of adding multi-direction feature fusion to the learned network. $\rightarrow$, $\uparrow$ and $\downarrow$ stand for horizontal, bottom-up and top-down feature fusion modules, respectively. Experiments are conducted on CVC \cite{cvc-db}.}
	{
		\begin{tabular}{ccccccc}
			\toprule
			Model & $\rightarrow$ & $\uparrow$ & $\downarrow$ & Jc (\%) &Params (M) &Flops (G)\\
			\midrule 
			U-Net \cite{unet} &\checkmark & $\times$ &    $\times$ &  82.29 &34.53 & 49.15\\
			\hline
			\multirow{4}*{\tabincell{c}{MixSearch \\($k$=2, $\mu$=0.5)}}  
			& \checkmark & & &  82.74 & 10.11 & 5.43 \\
			& \checkmark & \checkmark & &  83.20 & 12.27 & 8.24 \\
			& \checkmark & & \checkmark &  85.46 & 10.63 & 6.88 \\
			& \checkmark &\checkmark& \checkmark &  \textbf{87.00} & \textbf{12.79} &\textbf{9.69} \\			
			\bottomrule
		\end{tabular}
	}
	\label{feature_fusion}
\end{table}

\begin{figure}
	\centering
	\subfloat[]{\includegraphics[width=0.3\columnwidth]{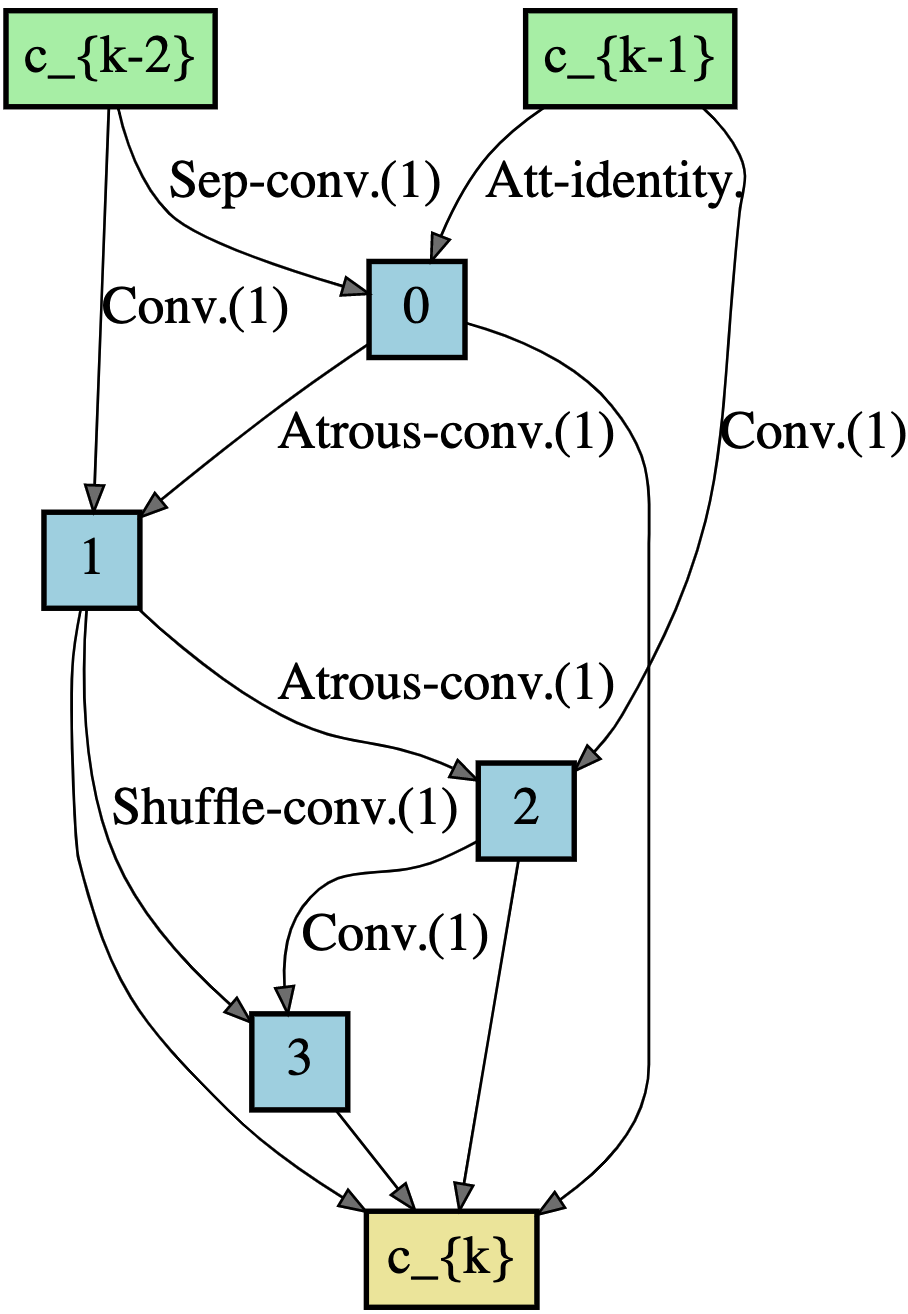}}%
	\subfloat[]{\includegraphics[width=0.45\columnwidth]{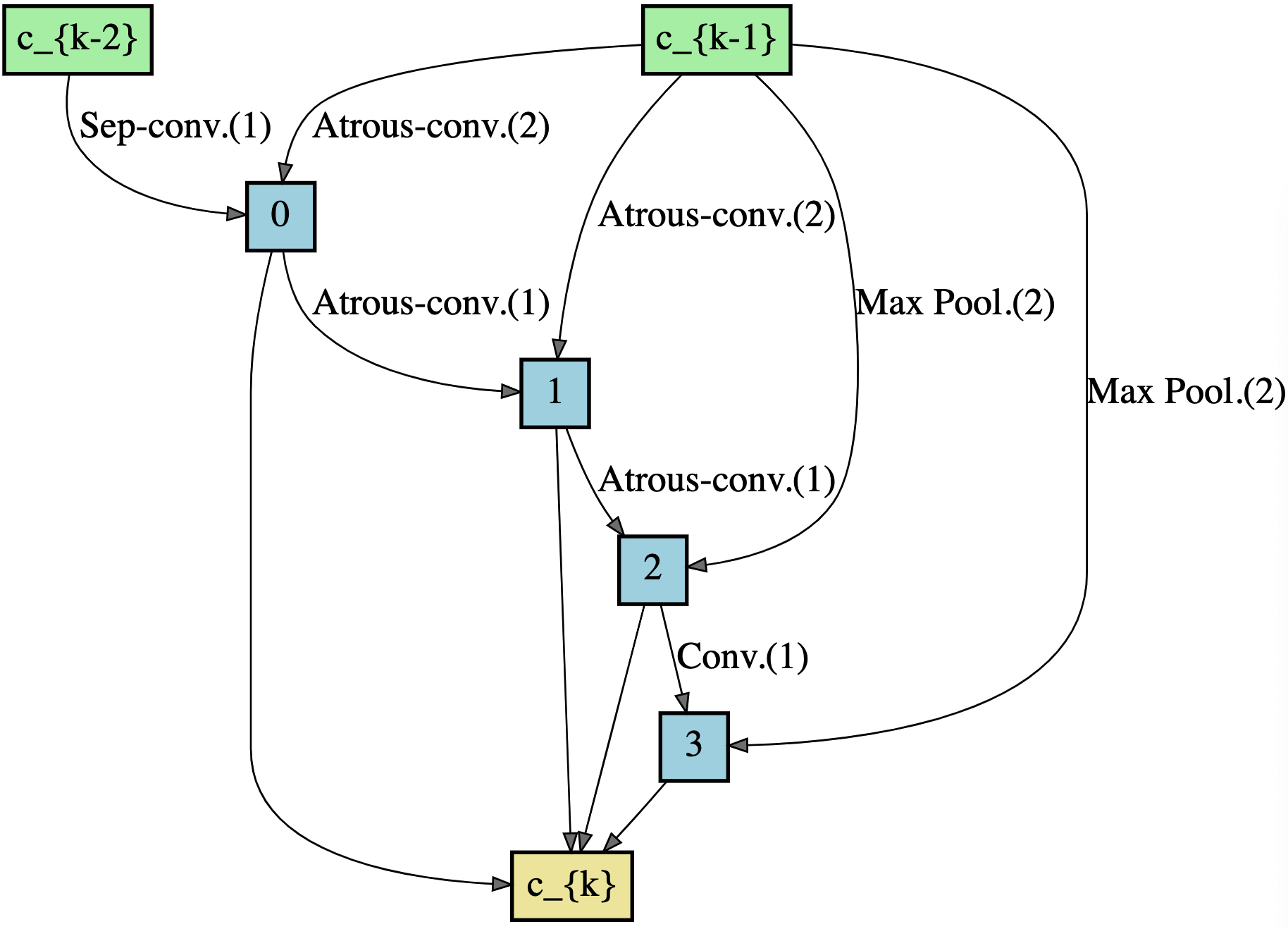}}%
	\subfloat[]{\includegraphics[width=0.25\columnwidth]{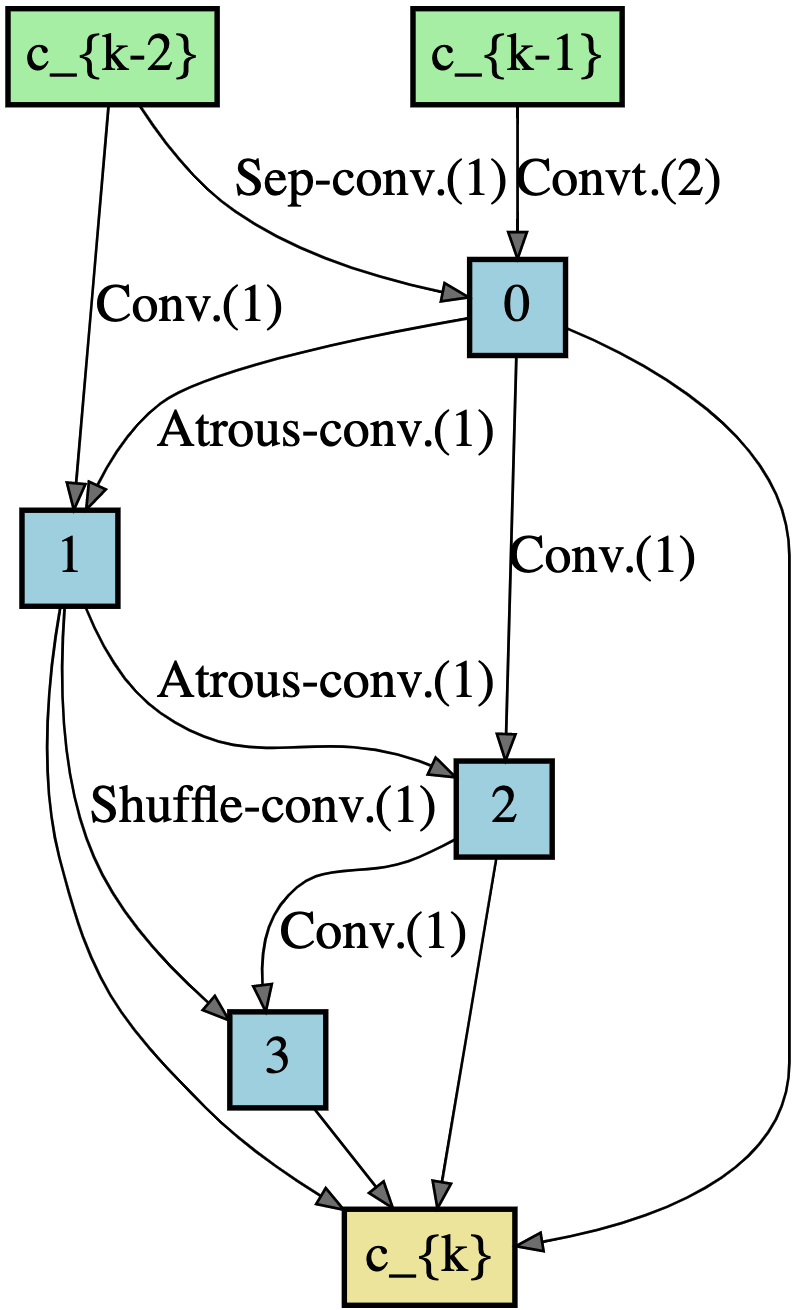}}%
	\caption{The architecture of different searchable cells learned by MixSearch. (a), (b) and (c) are the Normal cell, Downsample cell and Upsample cell architectures, respectively.}
	\label{fig:cell_archs}
\end{figure}

\subsection{Comparison with SOTA}
As we have mentioned above, MixSearch is based on a weaved encoder-decoder paradigm, which is similar to the U-Net. For comparison, we choose the U-Net \cite{unet} as the baseline model, while some other popular U-Net variants, such as U-Net++ \cite{unet++}, Attention U-Net \cite{attention_unet}, MR U-Net \cite{multiresunet} and R2T U-Net \cite{recurrent}, are also included. 

Under different search configurations, we report the experimental results in Table \ref{tab:compare}. Particularly, we introduce three settings to comprehensively evaluate the performance of various network architectures. The first setting, i.e., ``DedicatedSearch", is that we perform architecture search on each dataset mentioned above (i.e., ISIC, CVC and CHAOS-CT). Then, the learned network is evaluated on the corresponding dataset. The second setting, i.e., ``UnionSearch", is simply uniting the different datasets together and then perform the network architecture search. The last one is based on the proposed ``MixSearch" method, where we conduct architecture search on composited dataset and produce a \emph{single} network architecture. All the evaluation process is similar to the first setting, which is provided in Section IV.A. In practice, we employ different $k$ values (cf. Eq. (\ref{mixup1})) and report their performance with different $\mu$ choices, which control the mixing weights of $k$ images (cf. Eq. (3)). To demonstrate the strength of MixSearch, we also report the model size in Table \ref{tab:compare}. In the following, we analyze the experimental results based on the target datasets.

\begin{figure}
	\centering
	\subfloat[$V_{isic}$]{\includegraphics[width=0.45\columnwidth]{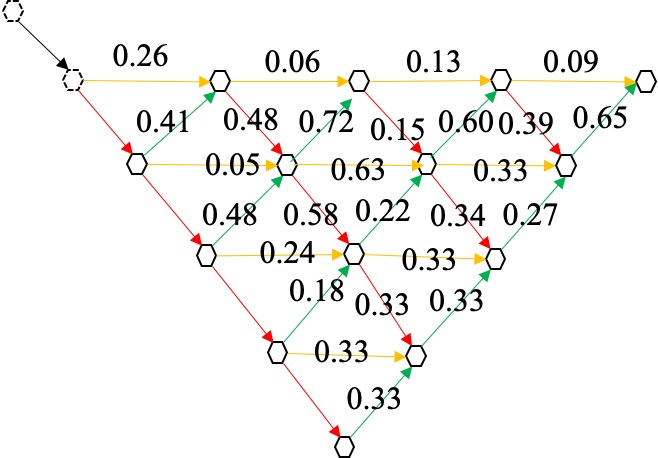}}%
	\quad
	\subfloat[$V_{mixup\_0.5}$]{\includegraphics[width=0.45\columnwidth]{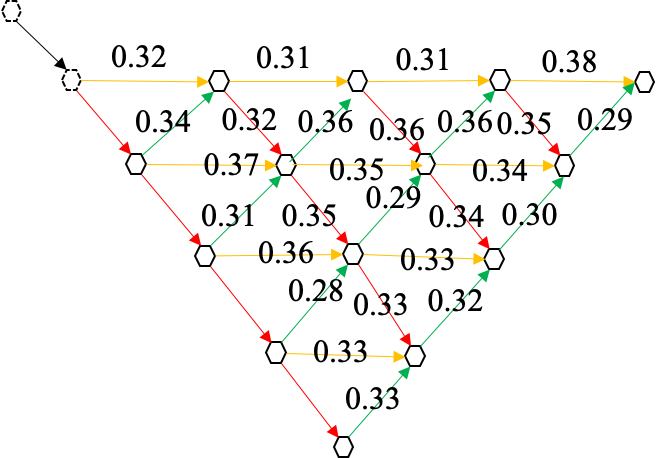}}%
	\caption{ The visualization of network architecture parameters $\beta$. (a) Optimal model learned by DedicatedSearch on dataset ISIC. (b) Optimal model learned by Mixsearch.}
	\label{fig:prune}
\end{figure}

\subsubsection{ISIC} For hand-crafted networks, Attention U-Net outperforms other baselines slightly (less than 1\% in both Dice and Jaccard scores). Unsurprisingly, among three architectures generated by DedicatedSearch, the one learned from ISIC produces the highest Dice score (88.53\%) and Jaccard score (79.8\%) on the target data of ISIC. When we conduct MixSearch on the composited dataset, the setting of \{$k$\shorteq 3, $\mu$\shorteq 1\} can achieve much better performance, surpassing the best hand-crafted architecture Attention U-Net by nearly 3\% in Dice score and over 4\% in Jaccard score. The evaluation results of UnionSearch are also shown in Table \ref{tab:compare}, from which we can see that although the learned network can produce comparable results with those of DedicatedSearch, our MixSearch can still surpass it by a large margin (90.07\% vs. 88.58\% and 82.68\% vs. 79.85\% in Dice and Jaccard scores, respectively).

\subsubsection{CVC} Similar to ISIC, in the DedicatedSearch, the architecture learned from its own dataset performs the best on the corresponding target data. Interestingly, we find that the architecture learned on CHAOS-CT performs the worst when ISIC and CVC are treated as target datasets. The underlying reason may be that there is a big domain gap between CHAOS-CT and the other two domains. Even so, MixSearch achieves the best performance again with \{$k$\shorteq 4, $\mu$\shorteq 0.5\}. 
In order to evaluate the generalization ability of the proposed method, we also implement the MixSearch on the randomly picked CVC (single) dataset. With the networks learned from MixSearch on CVC dataset, the segmentation performance outperforms that of DedicatedSearch and UnionSearch universally, which indicates that even on the single dataset, our proposed method can also learn more powerful and general network architectures. When learn from the mixed composited dataset consisted of various datasets, the performance and generalization ability of learned architectures can be further improved. 

\begin{figure}
	\centering
	\includegraphics[width=0.9\linewidth]{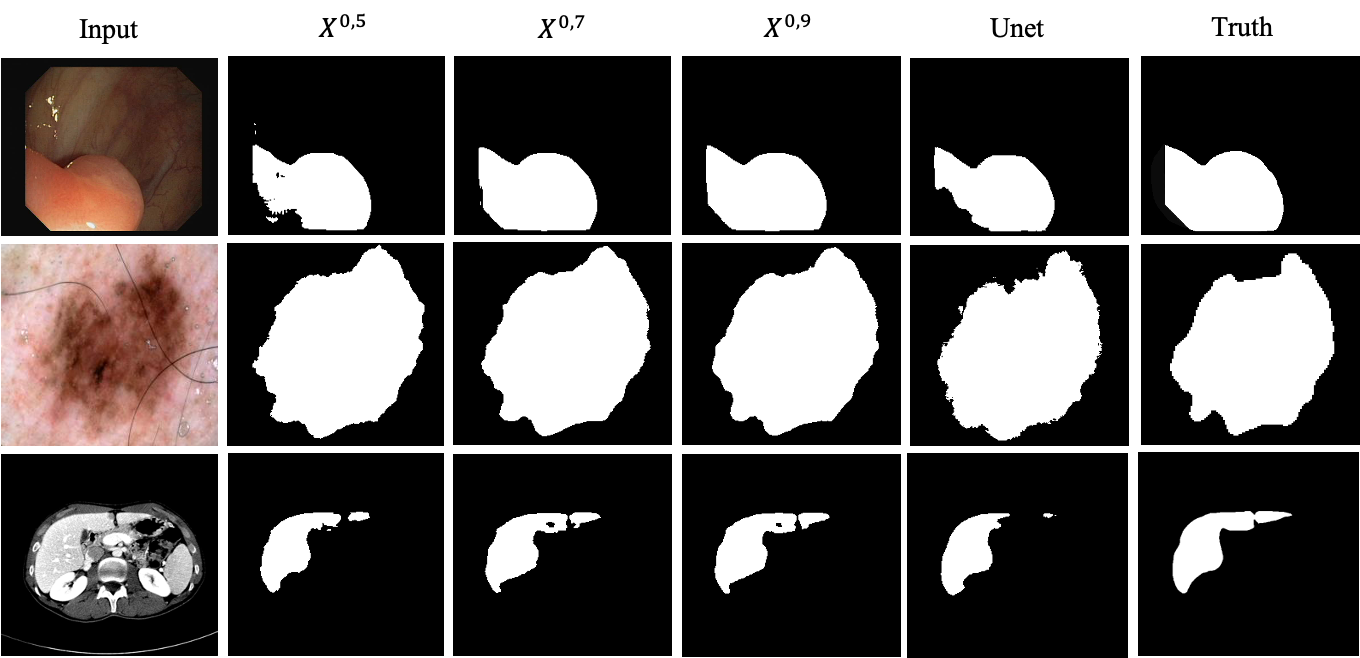}
	\caption{ Performance comparison between $V_{mixup\_0.5}$ and U-Net on the CVC, ISIC, and CHAOS-CT datasets (from top to bottom, respectively). $X^{0,5}$,$X^{0,7}$ and $X^{0,9}$ are three progressive prediction outputs of $V_{mixup\_0.5}$. }
	\label{fig:stepwise}
\end{figure}

\subsubsection{CHAOS-CT} As we have mentioned above, CHAOS-CT seems to be different from the other two datasets. We take CHAOS-CT as the source dataset and find that the basic U-Net achieves better results and outperforms recent state-of-the-art architectures by a slight margin. We guess the potential reason may be that the anatomical structures of the segmentation task on CHAOS-CT are simpler and much larger than those on ISIC and CVC datasets. Moreover, we find that the performance of networks learned by DedicatedSearch on CHAOS-CT is worse than those of U-Net and networks learned on CVC. Somewhat surprisingly, we observe that MixSearch can still produce promising results by filling the semantic gap between images from quite different domains.

The purpose of this work is learning domain generalized universal architectures for medical image segmentation, which are able to achieve satisfying performance on different datasets across different domains. Our proposed method is not restricted to any NAS-based algorithm and thus can also be easily applied on any of them. In order to further show the effectiveness of our proposed method, we take NAS-Unet \cite{nas-unet} as an example and extend our method on it to show the improved generalization ability of learned architectures. For fair comparison, we reproduce NAS-Unet using the same optimizer, learning rate, candidate operations and the same GPU memory footprint to our proposed method. Different to the proposed weaved search space (i.e., weaved space in Fig. \ref{fig:compare with NAS-UNET}), the NAS-Unet utilized the original U-Net architecture as search space, which is called plain search space (i.e., Plain-Space in Fig. \ref{fig:compare with NAS-UNET}) in this work. Fig. \ref{fig:compare with NAS-UNET} shows the performance of networks learned from plain and weaved search spaces, from which we can see that the performance of architectures learned from weaved search space outperforms that from the plain search space. 

We further evaluate NAS-Unet on the proposed mixed dataset, which is shown in Fig. \ref{fig:mixsearch of NAS-UNET}. We can see that, even though the DedicatedSearch using NAS-Unet can achieve outstanding performance on different datasets (i.e., ISIC and CHAO-CT), NAS-Unet with the proposed dataset mixing strategy can further improve the performance and generalization ability of learned architectures. It is worth noting that the performance of NAS-Unet with the mixed dataset on plain search space with $k=2$ reaches the peak value, and although the performance on $k=3$ and $k=4$ outperform the DedicateSearch and UnionSearch, they are worse than the performance on $k=2$. However, our proposed MixSearch with weaved search space can discover more powerful architectures on much more complex mixed datasets generated using $k=3$ and $k=4$. It is because that the plain search space used in NAS-Unet has a limited capacity to capture complex local and global image structures in the mixed dataset, which would further lead to lose useful feature characteristics. With our proposed weaved search space, the more accurate local and global features can be captured and fused together to provide discriminative and transferable information across different datasets and domains.

\begin{table}[htb]
	\centering
	\fontsize{7}{9}\selectfont    
	\caption{Jaccard scores of $X^{0,5}$, $X^{0,7}$, $X^{0,9}$ on the ISIC \cite{isic2018}, CVC \cite{cvc-db}, and CHAOS-CT \cite{chao2019} datasets..}
	\begin{tabular}{c|c|c|c|c|c|c}
		\toprule
		\multirow{2}{*}{Dataset} 
		&\multicolumn{2}{c|}{$X^{0,5}$} 
		&\multicolumn{2}{c|}{$X^{0,7}$} 
		&\multicolumn{2}{c}{$X^{0,9}$} \\
		\cline{2-7}
		\multirow{2}{*}& Dice (\%)& Jc (\%)&Dice (\%) &Jc (\%) &Dice (\%) &Jc (\%)\\
		\midrule
		ISIC &87.90 &78.79 & 88.44 & 79.63 &\textbf{88.55}& \textbf{79.86}   \\
		CVC      & 90.88&83.39 &92.71&86.50& \textbf{93.00} & \textbf{87.00}   \\
		CHAOS-CT &94.54& 89.79 &95.70& 91.88 &\textbf{95.79}& \textbf{92.03} \\
		\bottomrule
	\end{tabular}
	\label{tab:stepwise}
\end{table}

\begin{table*}[!htp]
	\centering
	\small
	\caption{Generalization ability of MixSearch and hand-crafted architectures. 
		\emph{DC} is channel doubling and \emph{DS} is deep supervision, respectively.}
	{
		\begin{tabular}{l|c|c|c|c|c|c|c|c|c}
			\toprule
			\multirow{2}{*}{Method} &\multirow{2}{*}{DC} &\multirow{2}{*}{DS}
			&\multicolumn{2}{c|}{ETIS} &\multicolumn{2}{c|}{LiTS} &\multicolumn{2}{c|}{KiTS} & Params \\
			\cline{4-9}
			& & & Dice (\%)& Jc (\%)&Dice (\%) &Jc (\%) &Dice (\%) &Jc (\%) & (M) \\
			\hline
			\hline
			U-Net \cite{unet} & \checkmark & & 91.32 & 84.12 & 95.19 & 91.02 &94.34&90.08 &34.53 \\
			U-Net++ \cite{unet++} & \checkmark & \checkmark & 90.75 & 83.28 & 95.34 & 91.24 &94.48 &90.27 & 36.63 \\
			Att. U-Net \cite{attention_unet} & \checkmark & & 91.75 & 84.90 & 95.26 & 91.14 & 94.28 &90.43 & 34.89\\
			MR U-Net \cite{multiresunet} & \checkmark & & 89.50& 81.30 & 95.35 & 91.30 & 94.26 &90.40 & 34.84\\
			R2T U-Net \cite{recurrent} & \checkmark & & 88.82& 80.31&92.36&86.55 &\textbf{94.64}&\textbf{90.53}& 39.09\\
			\hline
			UnionSearch & \checkmark & \checkmark & 91.05 & 83.64 & 95.04 & 90.70 &94.03&89.70& 9.71\\
			\hline
			\tabincell{c}{MixSearch\\ \{$k$\shorteq 4, $\mu$\shorteq 1\}} & \checkmark  & \checkmark & \textbf{92.00}&\textbf{85.37}&\textbf{95.71}&\textbf{91.88}&\textit{\textbf{94.53}}&\textit{\textbf{90.48}}&13.26 \\			
			\bottomrule
		\end{tabular}
	}
	\label{tab:compare_extra}
\end{table*}

\subsection{Effectiveness of Multi-direction Feature Fusion}
As we have mentioned before, the proposed search space contains top-down, horizontal and bottom-up operations to equalize resolution and standardize semantics. We argue that these operations also contribute to the performance improvements. In this section, we conduct an ablation study to demonstrate the influence of multi-direction feature fusion.

As shown in Table \ref{feature_fusion}, we verify the effectiveness of multi-direction feature fusion through ablation experiments on CVC. When only the horizontal feature fusion branch is retained, the searching strategy achieves comparable performance to U-Net while reducing the number of parameters by 3$\times$ and the amount of calculation (measured by FLOPS) by 8$\times$. This proves the potential of searchable cell in terms of improving segmentation performance and reducing the number of parameters. Further adding bottom-up feature fusion can achieve improvement of 2.7\%, while the model size and the amount of calculation are increased by 0.52 M and 1.45 GFLOPS, respectively. If choosing to add top-down feature fusion instead of bottom-up, the accuracy will increase by 0.46\%, and the model size and the computation cost increase by 2.16 M and 2.81 GFLOPS, respectively. Such comparison shows that both bottom-up and top-down feature fusion branches can improve model performance, but the former is more effective than the latter. When all three feature fusion branches are retained, the performance is improved by 4.3\%, which is greater than the sum of the benefits obtained by adding bottom-up and top-down branches, separately. These improvements verify that adding multi-direction feature fusion to the searching process can complement each other to boost the segmentation performance. 

\subsubsection{Network architecture analysis}
The architectures of learned Normal cell, Downsample cell and Upsample cell by MixSearch are shown in Fig. \ref{fig:cell_archs}. We can see from Fig. \ref{fig:cell_archs} that the searching process in our proposed weaved search space would select diverse operations rather than one certain types of operations. In addition, the standard skip connection (i.e., identity operation) is rarely selected in our experiments. It indicates that integrating multiple diverse operations may be able to capture ample information, including more accurate spatial and high-level semantic information, from complex mixed feature representations to help construct much more powerful network architectures with better generalization ability. 
In the network-level search, each intermediate feature map in the network contains top-down, horizontal and bottom-up feature fusion to equalize resolution and standardize semantics, while $\beta$ indicates the importance of each cell branch. We visualize the network architecture parameter $\beta$, which is shown in Fig. \ref{fig:prune}. $V_{isic}$ is the optimal model learned by DedicatedSearch on ISIC dataset, while $V_{mixup\_0.5}$ is the optimal model obtained by MixSearch with $k=2, \mu = 0.5$. 

\subsubsection{Stepwise feature aggregation}

The network learned by MixSearch contains rich feature aggregation, which can extract better semantic and spatial information. We further provide more experiments to prove it in this section.
The $X^{0,5}$, $X^{0,7}$, $X^{0,9}$ in Fig. \ref{fig:stepwise} are three outputs used for deep supervision in MixSearch, which are obtained by performing upsampling and $1 \times 1$ convolution on $X^{1,5}$, $X^{1,7}$ and $X^{1,9}$, respectively. From the visualization of segmentation masks, the output of $X^{0,5}$, $X^{0,7}$, $X^{0,9}$ become better and better, and the edges are more and more accurate. The quantitative evaluations of the segmentation results are shown in Table \ref{tab:stepwise}. As shown in Table \ref{tab:stepwise}, Jaccard score is getting higher with the increase of ``Layer".
The score of $X^{0,7}$ is significantly higher than that of $X^{0,5}$, while the score of $X^{0,9}$ is slightly higher than that of $X^{0,7}$. For applications with low resources (e.g., on a mobile device), if the segmentation accuracy meets the requirements, we can use  $X^{0,7}$ as the final output mask, which means that the entire decoding path of $X^{0,9}$ can be pruned, which would greatly reduce the amount of computation during inference.

\subsection{Generalization Ability of MixSearch}

In order to test the generalization ability of learned network architectures, we conduct experiments on three unseen datasets: ETIS \cite{silva2014toward}, LiTS \cite{lits} and KiTS \cite{heller2019state}, which come from different modalities and domains. ETIS is collected using wireless capsule endoscopy equipments, while LiTS and KiTS contain CT slices for the liver and kidney, respectively. We believe that it is a pretty challenging setting for the learned architectures because they have not seen similar images before and thus are not designed for them.
Table \ref{tab:compare_extra} displays the results on three unseen datasets. In ETIS, among the hand-crafted models, Attention U-Net achieves the best performance, while MR U-Net and R2T U-Net produce the best results in LiTS and KiTS, respectively. If we take into account the performance of all the three unseen datasets, Attention U-Net can be regarded as the most competitive hand-crafted network architecture. Meanwhile, it is very interesting to find that the basic U-Net only performs slightly worse than Attention U-Net, which demonstrates that the basic U-Net architecture can still achieve satisfying results in most cases with modern training strategies.

If we compare MixSearch with hand-crafted models, we can find that the learned architecture is able to surpass Attention U-Net on the three datasets by an obvious margin. Particularly, in the ETIS, MixSearch outperforms Attention U-Net by nearly 1\% in Dice score and 1.5\% in Jaccard similarity. Considering the results of best two hand-crafted architectures (Attention U-Net and U-Net) are very close, we argue the above improvements are quite impressive. In the LiTS, MR U-Net only surpasses the second best U-Net++ by 0.01\% in Dice score and 0.06\% in Jaccard score. In contrast, our MixSearch strategy outperforms MR U-Net by an obvious margin. In the KiTS, even though the R2T U-Net achieves the best performance and surpasses the other U-Net series by a large margin, MixSearch still achieves comparable performance with R2T U-Net by a narrow margin (94.53\% vs. 94.64\% and 90.48\% vs. 90.53\% in the Dice and Jaccard scores, respectively). 
Another interesting comparison in Table \ref{tab:compare_extra} is that UnionSearch performs worse than the basic U-Net architecture on all datasets. The underlying reason may be that the learned architectures might overfit the individual source datasets and thus generalize poorly to unseen datasets. However, the proposed MixSearch is able to produce generalizable and poweful architectures, since the domain gap between individual datasets is naturally extrapolated by cross-domain mixing.

\section{Conclusion}
In this paper, we proposed to construct a composited medical dataset in order to search for a high-performance and generalizable segmentation network. Specifically, we first presented a novel dataset mixing strategy to mix multi-domain medical datasets and construct composited medical images. Inspired by the U-shape network paradigm introduced by U-Net, we also built a weaved search space which contains built-in learnable horizontal and vertical feature fusion modules. This searching approach was then applied to the composited dataset to produce competitive segmentation architectures. Extensive experiments demonstrated the effectiveness of learned segmentation architectures in both intra-domain and inter-domain segmentation tasks.

\bibliographystyle{IEEEtran}
\bibliography{refs}

\end{document}